\documentclass[letterpaper, 10 pt, journal, twoside]{ieeetran} 
\usepackage{xspace}

\newcommand*{\ie}{i.e.\@\xspace}

\newcommand{\secref}[1]{Section \ref{#1}}

\newcommand{\tabref}[1]{Table \ref{#1}}

\newcommand{\figref}[1]{Figure \ref{#1}}

\usepackage[prependcaption,colorinlistoftodos]{todonotes}

\ifCLASSINFOpdf
\else
\fi
\usepackage{graphicx}
\graphicspath{{images/}}

\usepackage{amsfonts}

\usepackage{bm}

\usepackage{mathtools}

\usepackage{amsmath} 

\usepackage{gensymb}

\newcommand{\mdash}{\operatorname{\mathit -}}

\makeatletter
\newcommand*\bigcdot{\mathpalette\bigcdot@{.5}}
\newcommand*\bigcdot@[2]{\mathbin{\vcenter{\hbox{\scalebox{#2}{$\m@th#1\bullet$}}}}}
\makeatother

\usepackage{subfig}

\usepackage{hyperref}

\begin{document}
\title{Time-Optimal Trajectory Planning with Interaction with the Environment}

% \author{Michael~Shell,~\IEEEmembership{Member,~IEEE,}
%         John~Doe,~\IEEEmembership{Fellow,~OSA,}
%         and~Jane~Doe,~\IEEEmembership{Life~Fellow,~IEEE}% <-this % stops a space
% \thanks{M. Shell was with the Department
% of Electrical and Computer Engineering, Georgia Institute of Technology, Atlanta,
% GA, 30332 USA e-mail: (see http://www.michaelshell.org/contact.html).}% <-this % stops a space
% \thanks{J. Doe and J. Doe are with Anonymous University.}% <-this % stops a space
% \thanks{Manuscript received April 19, 2005; revised August 26, 2015.}}
\author{Vincenzo Petrone, Enrico Ferrentino, and Pasquale Chiacchio%
\thanks{Manuscript received: February, 24, 2022; Revised April, 24, 2022; Accepted June, 26, 2022.}%Use only for final RAL version
\thanks{This paper was recommended for publication by Editor Hanna Kurniawati upon evaluation of the Associate Editor and Reviewers' comments.} %Use only for final RAL version
\thanks{The authors are with Department of Computer Engineering, Electrical Engineering and Applied Mathematics (DIEM), University of Salerno, 84084 Fisciano, Italy
        (e-mail: {\tt\footnotesize vipetrone@unisa.it}; {\tt\footnotesize eferrentino@unisa.it}; {\tt\footnotesize pchiacchio@unisa.it})}%
\thanks{Digital Object Identifier (DOI): see top of this page.}
}

\markboth{Accepted version, published at https://doi.org/10.1109/LRA.2022.3191813 -- IEEE Robotics and Automation Letters}{}




\maketitle

\begin{abstract}
Optimal motion planning along prescribed paths can be solved with several techniques, but most of them do not take into account the wrenches exerted by the end-effector when in contact with the environment. When a dynamic model of the environment is not available, no consolidated methodology exists to consider the effect of the interaction. Regardless of the specific performance index to optimize, this article proposes a strategy to include external wrenches in the optimal planning algorithm, considering the task specifications. This procedure is instantiated for minimum-time trajectories and validated on a real robot performing an interaction task under admittance control. The results prove that the inclusion of end-effector wrenches affect the planned trajectory, in fact modifying the manipulator's dynamic capability.
\end{abstract}

% \begin{IEEEkeywords}
% IEEE, IEEEtran, journal, \LaTeX, paper, template.
% \end{IEEEkeywords}
\begin{IEEEkeywords}
Constrained Motion Planning, Optimization and Optimal Control, Compliance and Impedance Control
\end{IEEEkeywords}

% \fi
\IEEEpeerreviewmaketitle

\section{Introduction}

\IEEEPARstart{M}{otion} planning of robotic manipulators is the problem of determining a joint space trajectory, suitable for the robot controller to execute. The process is usually divided into two hierarchical sub-tasks. First, a geometric \emph{path} is defined in task or joint space, considering specific task requirements, environmental obstacles, joint mechanical limits. Then, a time law is assigned, resulting in a \emph{trajectory} respecting actuator constraints, such as maximum joint velocities and torques \cite{verscheure_time-optimal_2009}.

One common approach to time parametrization is to optimize some performance function, usually determined by the robot operational conditions. When the objective is to minimize the trajectory tracking time, the problem is usually referred to as \emph{Time-Optimal Trajectory Planning} (TOTP) \cite{ma_new_2021}.

\subsection{Motivation and Scope}

Considering motion planning with the aforementioned decoupled approach, the focus of this paper is on TOTP. The process input is a joint space path, possibly derived via inverse kinematics (IK), if the task is initially defined in task space.

The importance of minimum time trajectories has been widely discussed in the literature. In manufacturing, it is connected to the maximization of the plant throughput \cite{verscheure_time-optimal_2009}; in space, it might help optimizing the operations of assets whose usage is notably expensive \cite{ferrentino_evolutionary_2019}.

In practice, TOTP can be addressed with several strategies, a concise overview of which is reported in \cite{kaserer_nearly_2019}, and by modeling the robotic system and its constraints in many different ways. The resolution strategy depends on the selected model and the constraints of interest. Numerical integration approaches deal with time optimality through the explicit maximization of the path velocity \cite{kim_comparison_2005, dong_generalized_2005}. Convex optimization techniques \cite{verscheure_time-optimal_2009} address the problem from the same perspective, but viscous friction effects and limits on the trajectory's third-order derivatives cannot be included. Dynamic Programming (DP) can account for arbitrary forms of constraints and cost functions, but can be extremely slow \cite{shin_dynamic_1986, singh_optimal_1987}. Recently, a novel evolutionary approach based on genetic algorithms \cite{ferrentino_evolutionary_2019} has been proposed to specifically account for singular points.

The limitation of the state-of-the-art techniques is that they assume the manipulator's end-effector not in contact with the surrounding environment. This seldom is a realistic hypothesis in nowadays scenarios, since the robot certainly has to interact with the working environment to modify it. Welding, gluing and assembly are common examples of industrial tasks requiring interaction. If external contacts are not included in planning, but they do occur during the execution, the trajectory could be infeasible, i.e.\ the trajectory tracking performance could degrade. The reason of poor tracking lies in the wrong assumption made on the dynamic modeling of the system. If the torques exploited to exert wrenches on the environment are not considered in the planning phase, then the ones required to produce the motion might not be available at execution.

If the task is known a priori, and a dynamic characterization of the interaction can be devised, then the resulting wrenches can be explicitly included in planning \cite{kaserer_time_2020}, through a simple extension of the parametrized dynamic model originally proposed in the seminal works \cite{bobrow_time-optimal_1985} and \cite{shin_minimum-time_1985}. In this communication, the more realistic assumption is made that the interaction wrenches cannot be precisely modeled, but lower and upper limits are available from the task specification. Dynamic programming is preferred as the underlying resolution technique due to its flexibility \cite{shin_dynamic_1986}.

When it comes to trajectory execution, the problem of interacting with unknown environments can be addressed at control level with the employment of variable impedance controllers, with the environment being progressively modeled through learning \cite{li_efficient_2018}. This paper focuses on planning, leaving the actual success of the execution to the specific controller employed for trajectory tracking. Yet, some details about the employed controller, used to support the experiments on the UR10 robot, are discussed.

In view of the aforementioned considerations, in this paper, a modification of the state-of-the-art TOTP algorithm is proposed. In particular, by making the assumption that no information about the environment is available, only upper and lower bounds for interaction wrenches are included at planning level, as part of the task specification. To this aim, the equations provided to the TOTP solver are extended to account for these bounds. The planned trajectory is eventually executed on a real manipulator with an admittance controller, to allow the robot to interact with a blackboard.

\subsection{Overview}
The remainder of this paper is organized as follows.
\secref{sec:dynamic_modeling_of_robotic_arms} presents the dynamic model adopted in the planning process. \secref{sec:task_space_admittance_control_for_position_controlled_robots} describes the control scheme used in the trajectory execution phase. \secref{sec:time_optimal_trajectory_planning_with_interaction_with_the_enviornment} illustrates the proposed methodology, highlighting how interaction wrenches can be included in the planning algorithm. \secref{sec:experimental_results} reports the experimental validation, showing how external wrenches affect TOTP. In addition, the results of executing a time-optimal planned interaction trajectory on an admittance-controlled real manipulator are presented.

\section{Dynamic Modeling of Robotic Arms} \label{sec:dynamic_modeling_of_robotic_arms}

Consider a robotic manipulator with $n$ Degrees Of Freedom (DOF), being assigned a task of $m \le n$ DOF. Suppose the end-effector (EE) in contact with an (unknown) environment, such that the manipulator exerts some forces $\bm f_e$ and momenta $\bm m_e$, collected in the external wrench vector $\bm h_e \coloneqq \left( \bm f_e, \bm m_e \right)^T$.

Said $\bm q$ the $n \times 1$ vector of joint variables, the Equations Of Motion (EOM) relate the manipulator's generalized torques at the actuators $\bm \tau$ to the joint positions $\bm q(t)$ and the corresponding time derivatives, namely the joint velocities $\dot{\bm q}(t)$ and accelerations $\ddot{\bm q}(t)$. The EOM can be expressed in a matrix form with the following dynamic model:
\begin{equation} \label{eq:dynamic_model}
\bm B(\bm q) \ddot{\bm q} + \dot{\bm q}^T \bm C(\bm q) \dot{\bm q} + \bm F \dot{\bm q} + \bm g(\bm q) = \bm \tau - \bm J^T(\bm q) \bm h_e,
\end{equation}
where $\bm B(\bm q)$ is the $n \times n$ inertia matrix; $\bm C(\bm q)$ is the $n \times n \times n$ centrifugal and Coriolis effects matrix; $\bm F$ is the diagonal $n \times n$ friction matrix; $\bm J^T(\bm q) \bm h_e$ are the torques needed to exert the wrenches $\bm h_e$ at the EE, expressed in the base frame, with $\bm J(\bm q)$ the manipulator's Jacobian.

It is worth remarking that \eqref{eq:dynamic_model} considers a model for friction: since high-speed trajectories are supposed to be commanded to the manipulator, a considerable portion of actuation capabilities is exploited in compensating this component. Thus, the term cannot be ignored in the model for TOTP, otherwise non-negligible inaccuracies are introduced in the torques computation.

\section{Task-Space Admittance Control for Position-Controlled Robots} \label{sec:task_space_admittance_control_for_position_controlled_robots}

The feasibility of a wide number of practical tasks relies on the control of the physical interaction between the robot and the EE, as pure motion control is possible only if an accurate planning of the task is available, requiring an exact characterization of the environment's geometrical and mechanical features. Since this hypothesis is unlikely to hold, interaction control has been largely studied in the last decades \cite{siciliano_springer_2016}.

Interaction controllers are employed to control the manipulator motion while considering the presence of wrenches arising as an effect of the interaction. Impedance and admittance control are two popular approaches, being the latter more suitable for position-controlled robots. In fact, given a certain force error as input, it produces a corresponding displacement as output, according to a mechanical admittance. Thus, when a desired task-space trajectory $\bm x_d(t)$ is received, the control action modifies the position reference, producing a compliant trajectory $\bm x_c(t)$ to be commanded to the robot. This is of utmost importance as industrial robots typically expose a position interface \cite{li_efficient_2018}.

Suppose that, along with a task space desired trajectory $\bm x_d(t)$, a reference wrench $\bm h_d(t)$ is given, too. Denote with $\bm \tilde{\bm h} \coloneqq \bm h_d - \bm h_e$ the wrench error and with $\tilde{\bm z} \coloneqq \bm x_d - \bm x_c$ the admittance control action, namely the deviation from the desired reference to the compliant task space command. The admittance control imposes mass-spring-damper EE dynamics, described by
\begin{equation} \label{eq:admittance_control}
\bm M_d \ddot{\tilde{\bm z}} + \bm K_D \dot{\tilde{\bm z}} + \bm K_P \tilde{\bm z} = - \tilde{\bm h},
\end{equation} 
where $\bm M_d$, $\bm K_D$ and $\bm K_P$ are $6 \times 6$ diagonal matrices corresponding, respectively, to mass, damping and stiffness matrices. The coefficients in the matrices are the control parameters, to be suitably selected depending on the desired interaction dynamics.

\begin{figure}[!b]
\centering
\includegraphics[width=\columnwidth]{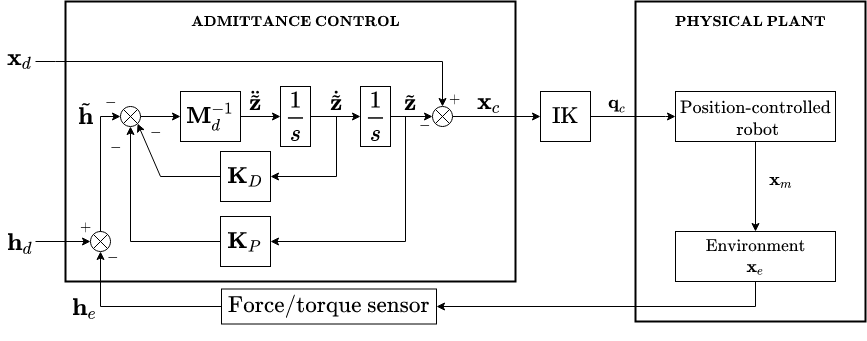}
\caption{Admittance control block scheme for position-controlled robots}
\label{fig:AdmittanceControl}
\end{figure}

From \eqref{eq:admittance_control}, the task space compliant frame is computed:
\begin{subequations}\label{eq:admittance_control_compliant_frame}
\begin{align}
\ddot{\tilde{\bm z}} &= \bm M_d^{-1} ( - \tilde{\bm h} - \bm K_D \dot{\tilde{\bm z}} - \bm K_P \tilde{\bm z} ), \\
\dot{\tilde{\bm z}} &= \int_{t} \ddot{\tilde{\bm z}} \,dt, \\
\tilde{\bm z} &= \int_{t} \dot{\tilde{\bm z}} \,dt, \\
\bm x_c &= \bm x_d - \tilde{\bm z}.
\end{align}
\end{subequations}

Finally, the joint-space command $\bm q_c$ is obtained by IK from $\bm x_c$, and set as the reference of the inner position loop, usually provided by the robot's manufacturer. If in contact with the environment, the resulting EE pose $\bm x_m$ produces the wrenches $\bm h_e$, as a result of the environment rest position $\bm x_e$ and the corresponding virtual penetration $\Delta \bm x \coloneqq \bm x_e - \bm x_m$.

Figure \ref{fig:AdmittanceControl} shows the block scheme corresponding to equations \eqref{eq:admittance_control} and \eqref{eq:admittance_control_compliant_frame}, highlighting how the wrench $\bm h_e$ is fed back to the control loop. To this aim, a force/torque sensor is implied in this work, but other solutions are possible to estimate the EE wrench, even without additional sensing \cite{kaserer_admittance_2016, wahrburg_motor-current-based_2018}.

\section{Time Optimal Trajectory Planning with Interaction with the Environment} \label{sec:time_optimal_trajectory_planning_with_interaction_with_the_enviornment}

In general, trajectory planning deals with assigning a time law to a pre-designed joint space path $\bm q(\lambda)$ or task space path $\bm x(\lambda)$, where $\lambda$ is a curvilinear coordinate describing the path geometry as a monotonic function of time, \ie
\begin{equation}
\lambda(t): \mathbb{R} \rightarrow [0;\Lambda], \dot{\lambda}=\frac{d\lambda}{dt} > 0 \; \forall t,
\end{equation}
with $\Lambda$ being the total path length. The relationship $\lambda(t)$ is the time law to find as a result of the time parametrization process, which is usually addressed through optimization. Once the optimal time law is found, the path can be time-parametrized, obtaining the joint space trajectory $\bm q \left( \lambda(t) \right) = \bm q(t)$. 

In this communication, TOTP is considered, meaning that $\lambda(t)$ is computed so as to minimize the overall duration of the task, considering robot-specific architectural limits and other application-dependent constraints. Among them, the inclusion of external interaction wrenches is considered.

\subsection{Dynamic model parametrization} \label{sec:problem_formulation}

Suppose that a task space path $\bm x(\lambda)$ is available from the task specification. Suppose to plan for a non-redundant robot, thus a joint space path $\bm q(\lambda)$ can be easily computed via IK.

Denote with $\bm q'(\lambda)$ and $\bm q''(\lambda)$ the first- and second-order derivatives of $\bm q(\lambda)$ with respect to $\lambda$. The time derivatives are obtained by simply applying the chain rule:
\begin{align}
\dot{\bm q} &= \bm q'(\lambda) \dot \lambda  \label{eq:joint_velocity}, \\
\ddot{\bm q} &= \bm q'(\lambda) \ddot{\lambda} + \bm q''(\lambda) \dot{\lambda}^2. \label{eq:joint_acceleration}
\end{align}

Plugging \eqref{eq:joint_velocity} and \eqref{eq:joint_acceleration} into \eqref{eq:dynamic_model} yields
\begin{equation} \label{eq:dynamic_model_lambda}
\bm a(\lambda) \ddot{\lambda} + \bm b(\lambda) \dot{\lambda}^2 + \bm c(\lambda) \dot \lambda + \bm g(\lambda) = \bm \tau - \bm J^T(\lambda) \bm h_e (\lambda),
\end{equation}
where
\begin{align}
	\bm a(\lambda) &= \bm B\big(\bm q(\lambda)\big) \bm q'(\lambda), \label{eq:a_lambda}\\
	\bm b(\lambda) &= \bm B\big(\bm q(\lambda)\big) \bm q''(\lambda) + \big(\bm q'(\lambda) \big)^T \bm C\big(\bm q(\lambda)\big) \bm q'(\lambda), \label{eq:b_lambda}\\
	\bm c(\lambda) &= \bm F \bm q'(\lambda), \label{eq:c_lambda}\\
	\bm g(\lambda) &= \bm g\big(\bm q(\lambda)\big), \label{eq:g_lambda}\\
	\bm J(\lambda) &= \bm J\big(\bm q(\lambda)\big). \label{eq:j_lambda}
\end{align}

As regards the wrench $\bm h_e$, a $\lambda$-parametrization is possible only if a dynamic characterization of the environment is available. For instance, \cite{kaserer_time_2020} proposes a dynamic model for a manipulated object, and assumes the external wrench to result from the object's planned motion. When the interaction model is not known, there is no particular strategy to plan for external wrenches. However, they can be included with the meaning of expected maximum and minimum exerted wrenches, given the task specification.

\subsection{Constraints parametrization} \label{sec:constraints_parametrization}

The TOTP problem is subject to constraints, given by the manipulator structure and the specific task. It is useful to define such limits with lower and upper bounds on a constrained variable, denoted with $(\underline{\bullet})$ and $(\overline{\bullet})$ respectively.

The robot inherent technological limits must be included in planning, so as to reach a feasible trajectory in terms of the actual manipulator capabilities. Typically, the joint actuators are limited in both torques and velocities. For each $j \in \lbrace 0, \ldots, n \rbrace$, the $j$-th joint's maximum and minimum torque and velocity limits are respectively expressed as
\begin{align}
\underline{\tau}_j & \le \tau_j \le \overline{\tau}_j, \label{eq:torque_limits}\\
\underline{\dot{q}}_j & \le \dot{q}_j \le \overline{\dot{q}}_j. \label{eq:joint_velocity_limits}
\end{align}

An exact characterization of the robot-environment interaction is hard to devise. More commonly, depending on the task specification, it is possible to define lower and upper bounds on the external wrenches vector. Typically, they correspond to the maximum and minimum wrenches that the EE and the environment can tolerate, or that are required by the task to achieve a certain level of quality. These can be seen as application-dependent constraints:
\begin{equation} \label{eq:wrench_limits}
\underline{\bm h_e}(\lambda) \le \bm h_e \le \overline{\bm h_e}(\lambda).
\end{equation}
As the dependency on $\lambda$ suggests, it is worth remarking that such limits do not need to be constant, but can be adapted along the path.

The task specifications, including the maximum stress the working environment must be subject to, are usually detailed in industrial contexts, whilst are typically not well-defined in service tasks. In such cases, the bounds could be identified by performing preliminary experiments and recording the measured wrenches, thus deciding upon the limits to enforce according to a certain degree of safety or quality of the executed task. If it is not possible to replicate the real-case scenario in which the planned trajectory has to be executed, and a physical copy of the system is not available, then at least a high-fidelity simulated environment should be adopted for that purpose.

One typical choice for the limit wrenches for common interaction tasks is when the contact only concerns one direction. For simplicity, it can be made to correspond with a specific axis of the base frame, for instance
\begin{equation*} \label{eq:single_contact_wrench}
\begin{split}
\underline{\bm h_e}(\lambda) &= \big(\underline f_N (\lambda), \underline f_{T_1}(\lambda), \underline f_{T_2}(\lambda), 0, 0, 0\big)^T, \\
\overline{\bm h_e}(\lambda) &= \big(\overline f_N (\lambda), \overline f_{T_1}(\lambda), \overline f_{T_2}(\lambda), 0, 0, 0\big)^T,
\end{split}
\end{equation*}
where $f_N(\lambda)$ is the normal force along the contact direction (which is supposed to be the $x$ axis in the base frame), and $f_{T_1}(\lambda)$ and $f_{T_2}(\lambda)$ are the tangential force components along the orthogonal motion directions, represented by the unit vector $\bm t = \big(t_1, t_2\big)^T$, defined in the tangential plane. The maximum and minimum tangential force components relate to the normal force according to
\begin{equation*} \label{eq:tan_force_components}
\begin{split}
\underline f_{T_1}(\lambda) & = \mu \underline f_N (\lambda) t_1 (\lambda) \quad \overline f_{T_1}(\lambda) = \mu \overline f_N (\lambda) t_1 (\lambda) \\
\underline f_{T_2}(\lambda) & = \mu \underline f_N (\lambda) t_2 (\lambda) \quad \overline f_{T_2}(\lambda) = \mu \overline f_N (\lambda) t_2 (\lambda)
\end{split}
\end{equation*}
where $\mu > 0$ is the friction coefficient. It can be available by knowledge of the EE and environment surfaces or easily estimated from trials, as in the experimental case of Section \ref{sec:experimental_results}. Clearly, the tangential forces can, in general, behave according to a more complicated (or, as in the worst case, completely unknown) model. If a model is supposed for the tangential components, its parameters must be estimated, and the model itself must be validated during preliminary experiments. Otherwise, tangential forces can be given, as for normal forces, in terms of independent upper and lower bounds that are unrelated to the normal force ones. The methodology we propose does not restrict the way tangential forces bounds are devised.

Denote with $\gamma_j(\lambda) \coloneqq \bm J_j^T(\lambda) \bm h_e (\lambda)$ the torque to apply to joint $j$ to exert wrenches $\bm h_e$ on the environment in a specific configuration $\bm q(\lambda)$. Then, starting from \eqref{eq:torque_limits}, the following inequality holds
\begin{equation}
\underline{\tau_j} - \gamma_j(\lambda) \le \tau_j - \gamma_j(\lambda) \le \overline{\tau_j} - \gamma_j(\lambda).
\end{equation}

Because of \eqref{eq:wrench_limits}, $\gamma_j$ can vary in the range
\begin{equation} \label{eq:gamma_limits}
\bm J_j^T(\lambda) \underline{\bm h_e}(\lambda) \le \gamma_j(\lambda) \le \bm J_j^T(\lambda) \overline{\bm h_e} (\lambda),
\end{equation}
hence $\tau_j - \gamma_j$ must respect the strictest limit given by \eqref{eq:gamma_limits}, \ie
\begin{equation} \label{eq:tau_gamma_limits}
\begin{split}
\max\big\{& \underline{\tau_j} - \underline{\gamma_j}(\lambda), \underline{\tau_j} - \overline{\gamma_j}(\lambda)\big\}
\le \tau_j - \gamma_j(\lambda) \\
\le \min\big\{& \overline{\tau_j} - \underline{\gamma_j}(\lambda), \overline{\tau_j} - \overline{\gamma_j}(\lambda)\big\},
\end{split}
\end{equation}
where
\begin{align}
\underline{\gamma_j}(\lambda) &= \bm J_j^T(\lambda) \underline{\bm h_e} (\lambda), \\
\overline{\gamma_j}(\lambda) &= \bm J_j^T(\lambda) \overline{\bm h_e} (\lambda).
\end{align}
By taking $\underline{\tau_j}$ and $\overline{\tau_j}$ out of the $\min$ and $\max$ operators, \eqref{eq:tau_gamma_limits} becomes
\begin{equation*} 
\begin{split}
\underline{\tau_j} + \max\big\{- & \underline{\gamma_j}(\lambda), - \overline{\gamma_j}(\lambda)\big\}
\le \tau_j - \gamma_j(\lambda) \\
\le \overline{\tau_j} + \min\big\{- & \underline{\gamma_j}(\lambda), - \overline{\gamma_j}\big\},
\end{split}
\end{equation*}
from which it follows that
\begin{equation} \label{eq:tau_h_limits}
\begin{split}
\underline{\tau_j} - \min\big\{& \underline{\gamma_j}(\lambda),\overline{\gamma_j}(\lambda)\big\}
\le \tau_j - \gamma_j(\lambda) \\
\le \overline{\tau_j} - \max\big\{& \underline{\gamma_j}(\lambda),\overline{\gamma_j}(\lambda)\big\}.
\end{split}
\end{equation}

Plugging \eqref{eq:dynamic_model_lambda} into the constraint \eqref{eq:tau_h_limits} yields
\begin{equation} \label{eq:dynamic_model_lambda_limited}
\begin{split}
\underline \tau_j - \underline{\tau_h}_j(\lambda) & \le a_j(\lambda) \ddot{\lambda} + b_j(\lambda) \dot{\lambda}^2 + c_j(\lambda) \dot{\lambda} + g_j(\lambda) \\
& \le \overline \tau_j - \overline{\tau_h}_j(\lambda),
\end{split}
\end{equation}
where
\begin{align}
\underline{\tau_h}_j(\lambda) &= \min\big\{\underline{\gamma_j}(\lambda),\overline{\gamma_j}(\lambda)\big\}, \\
\overline{\tau_h}_j(\lambda) &= \max\big\{\underline{\gamma_j}(\lambda),\overline{\gamma_j}(\lambda)\big\}.
\end{align}

Solving inequality \eqref{eq:dynamic_model_lambda_limited} for $\ddot{\lambda}$ and accounting for the strictest limit among all $n$ joints, the following constraint is obtained:
\begin{equation} \label{eq:l_lambdadd_u}
L(\lambda, \dot \lambda) \le \ddot{\lambda} \le U(\lambda, \dot \lambda),
\end{equation}
where
\begin{align}
L(\lambda, \dot \lambda) = \max_{j=1,\ldots,n} L_j(\lambda, \dot \lambda), \\
U(\lambda, \dot \lambda) = \min_{j=1,\ldots,n} U_j(\lambda, \dot \lambda),
\end{align}
with
\begin{align}
L_j(\lambda, \dot \lambda) &= \frac{
\splitfrac{
	\splitfrac{
		\big(\underline \tau_j - \underline{\tau_h}_j(\lambda) \big) \delta_j(\lambda) +
	}{
		+ \big(\overline \tau_j - \overline{\tau_h}_j(\lambda) \big) \big(1 - \delta_j(\lambda) \big) +
	}
}{
-
b_j(\lambda)\dot{\lambda}^2 - c_j(\lambda)\dot \lambda - g_j(\lambda)
}}{
a_j(\lambda)
}, \label{eq:l_j} \\
U_j(\lambda, \dot \lambda) &= \frac{
\splitfrac{
	\splitfrac{
		\big(\overline \tau_j - \overline{\tau_h}_j(\lambda) \big) \delta_j(\lambda) +	
	}{
		+ \big(\underline \tau_j - \underline{\tau_h}_j(\lambda) \big) \big(1 - \delta_j(\lambda) \big) +	
	}
}{-
b_j(\lambda)\dot{\lambda}^2 - c_j(\lambda)\dot \lambda - g_j(\lambda)
}}{
a_j(\lambda)
} \label{eq:u_j}
\end{align}
and
\begin{equation}
{ \delta_j(\lambda) } = 
\begin{cases}
1, & {\text{if}}\ a_j(\lambda) > 0 \\
{0,} & {\text{if}}\ a_j(\lambda) < 0
\end{cases}.
\end{equation}

While \eqref{eq:l_lambdadd_u} bounds $\ddot \lambda$, \eqref{eq:joint_velocity_limits} produces a limit on $\dot \lambda$, since joint velocities are a function of the pseudo-velocity, as evident from \eqref{eq:joint_velocity}:
\begin{equation} \label{eq:lambda_dot_limits}
\max_{j=1, \ldots, n}{\frac{ \underline{\dot q}_j }{ q_j'(\lambda) } } \le \dot \lambda \le \min_{j=1, \ldots, n}{\frac{ \overline{\dot q}_j }{ q_j'(\lambda) } }.
\end{equation}

\subsection{Solution using a Dynamic Programming algorithm} \label{sec:solution_using_a_dynamic_programmin_algorithm}

The objective of the TOTP algorithm is to minimize the overall trajectory tracking time. Let $\bm s(t) = \big[\lambda(t), \dot \lambda(t)\big]^T$ be the \emph{state} and $\ddot \lambda(t)$ the \emph{control input} of the considered dynamical system. The problem of finding the minimum-time trajectory can be stated as follows:
\begin{equation} \label{eq:t_opt}
\min_{\ddot \lambda(t) \in \left[L(\bm s),U(\bm s)\right]}{\int_{0}^{t_f}{\,dt}},
\end{equation}
subject to \eqref{eq:lambda_dot_limits} and the following boundary conditions:
\begin{equation} \label{eq:boundary_conditions}
\begin{split}
{\bf s}(0) & = \left[ 0, 0 \right]^T, \\
{\bf s}(t_f) & = \left [\Lambda, 0 \right]^T,
\end{split}
\end{equation}
ensuring that the robot will start and stop at rest.

In this paper, the same DP algorithm as in \cite{shin_dynamic_1986} is adopted to solve the TOTP problem above, \ie the phase plane $(\lambda \mdash \dot \lambda)$ is discretized to get a rectangular grid of $N_\lambda \times N_{\dot \lambda}$ cells. The cost function uses the approximation proposed in \cite{singh_optimal_1987} for the time discretization from one cell to another.

By employing such a framework, valid nodes for the solution are selected that satisfy conditions \eqref{eq:l_lambdadd_u}, \eqref{eq:lambda_dot_limits} and \eqref{eq:boundary_conditions} for each node in the grid. After the trajectory is planned, the parametric velocities and accelerations are turned into joint velocities and accelerations according to \eqref{eq:joint_velocity} and \eqref{eq:joint_acceleration}.

The reader should be aware that DP is not the only technique suitable for implementing a TOTP algorithm, but the choice of the dynamic model \eqref{eq:dynamic_model}, with the inclusion of viscous friction effects, and the corresponding $\lambda$-parametrization \eqref{eq:dynamic_model_lambda}, breaks the limitations of the convex optimization approach \cite{verscheure_time-optimal_2009}, as also pointed out in \cite{ferrentino_evolutionary_2019, kaserer_nearly_2019}.

Also, it is worth remarking that, since the proposed methodology only replaces the limits to be imposed in the optimization process, it can be employed to optimize the time law with respect to any desirable performance index.

\section{Experimental Results} \label{sec:experimental_results}

\begin{figure}[!t]
\centering
\includegraphics[width=\columnwidth]{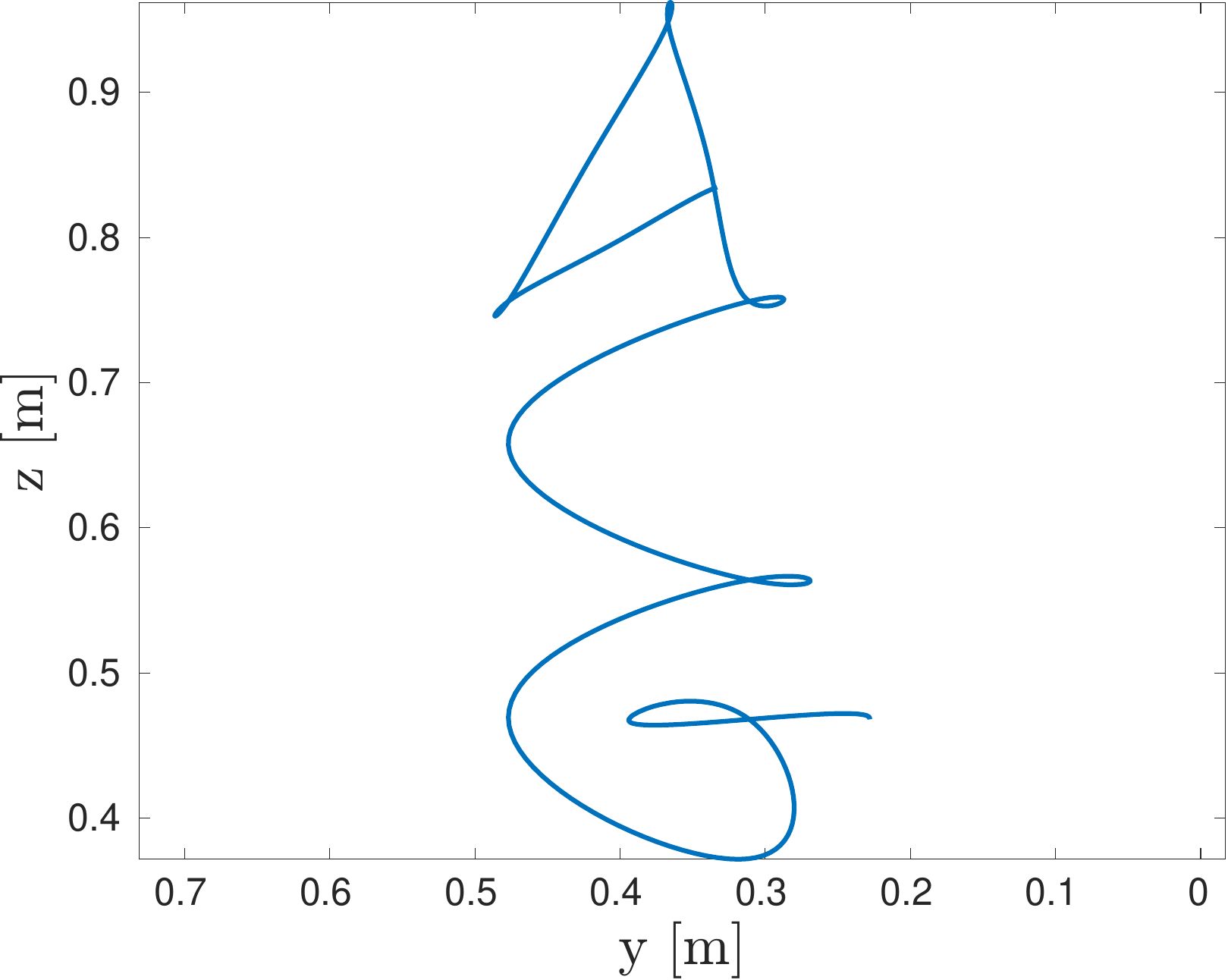}
\caption{The path to time-parametrize, represented on the writing plane, \ie $(y,z)$ in the base frame}
\label{fig:ACG}
\end{figure}

The interaction task is a blackboard writing task, \ie the 6-DOF UR10 manipulator is required to write the Automatic Control Group (ACG) acronym (illustrated in \figref{fig:ACG}) on a blackboard using some chalk. The EE is made of a piece of chalk inserted in a properly 3D modeled and printed mechanical interface.

\begin{figure}[!b]
\centering
\includegraphics[width=\columnwidth]{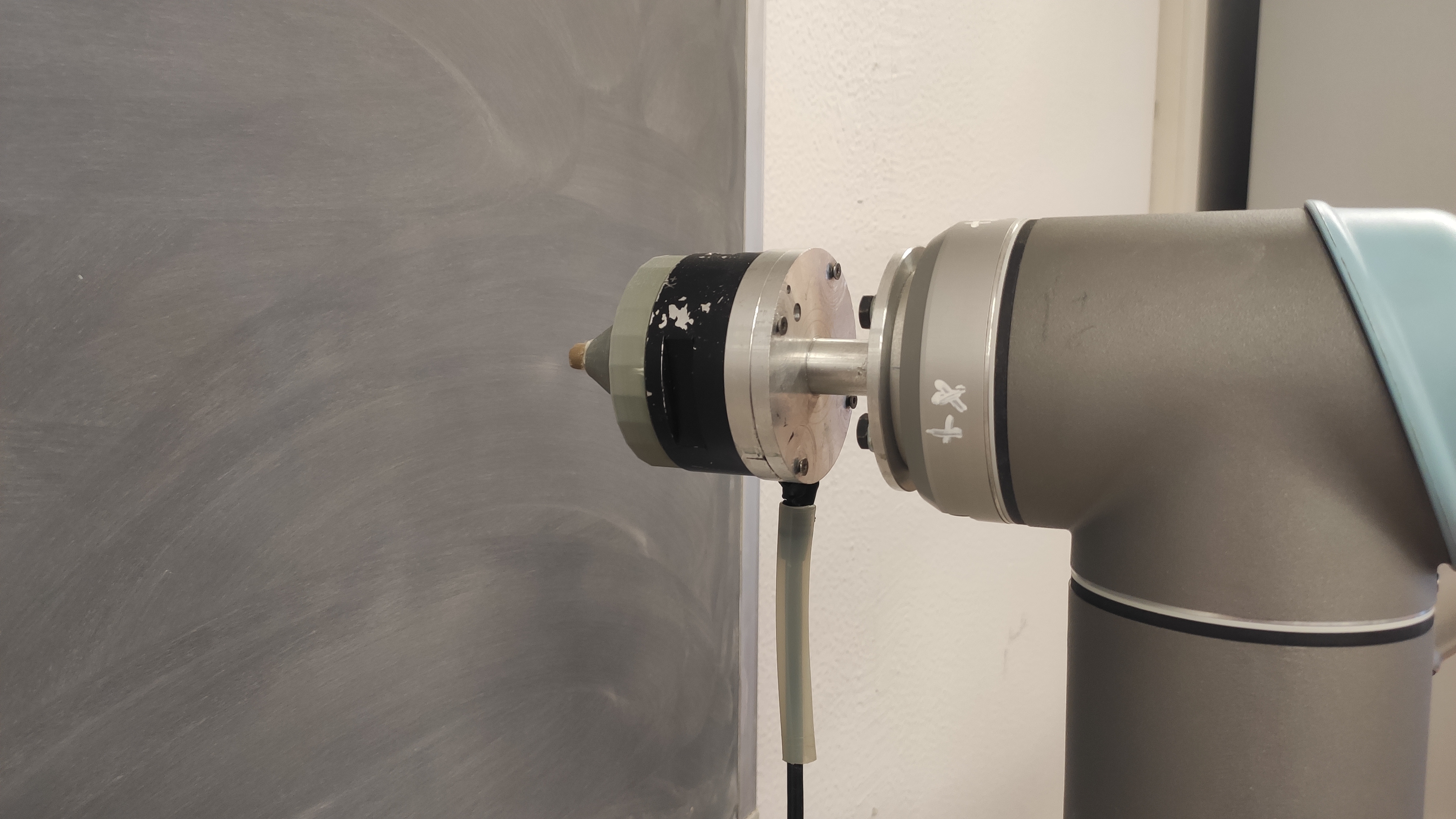}
\caption{Experimental setup: the UR10 robot is equipped with a force/torque sensor, on top of which a mechanical interface is mounted, with a piece of chalk inserted in it}
\label{fig:experimental_setup}
\end{figure}

The experimental setup is shown in \figref{fig:experimental_setup}. With reference to the base frame, the contact direction is the $x$ axis, while the free motion directions (that define the ``writing plane'') are the $y$ and $z$ axes. The wrenches are sensed via a flange-mounted 6-DOF force/torque sensor. The chalk can tolerate up to $\overline f_N=80 \, \text{N}$, while a clean stroke is guaranteed with $\underline f_N=1 \, \text{N}$. The friction coefficient $\mu = 0.519$ has been experimentally estimated from force measurements. The torque and velocity limits at the manipulator's joints are reported in \tabref{tab:torque_velocity_limits}.

As regards the planning process, the inequalities \eqref{eq:l_lambdadd_u} are verified by evaluating the vectors \eqref{eq:a_lambda}--\eqref{eq:g_lambda} for each $\lambda$, employing the UR10's dynamic model identified in \cite{gaz_model-based_2018}.

The admittance controller uses the parameters $\bm M_d = \text{diag}\{0.1, 0.1, 0.02\}$, $\bm K_D = \text{diag}\{300, 300, 1200\}$ and $\bm K_P = \text{diag}\{5500, 5500, 625\}$ so as to be compliant and damped along the contact axis and stiff and less damped along the motion/writing plane. The controller receives a force reference of $f_d = 20$ N in order to ensure a stable contact. 

\begin{table}[!t]
\caption{Torque and velocity limits used for TOTP}
\label{tab:torque_velocity_limits}
\begin{center}
\begin{tabular}{c|c|c|c}
axis & joint name & $\overline{\dot{\bm q}}=-\underline{\dot{\bm q}} \; [\degree / \text s]$ & $\overline{\bm \tau} = - \underline{\bm \tau} \; [\text{Nm}]$  \\ 
\hline
1    & base       & 40                                  & 80                            \\
2    & shoulder   & 40                                  & 80                            \\
3    & elbow      & 60                                  & 100                           \\
4    & wrist 1    & 60                                  & 28                            \\
5    & wrist 2    & 60                                  & 28                            \\
6    & wrist 3    & 60                                  & 28                            \\
\hline
\end{tabular}
\end{center}
\end{table}

\begin{figure}[!b]
\centering
\includegraphics[width=\columnwidth]{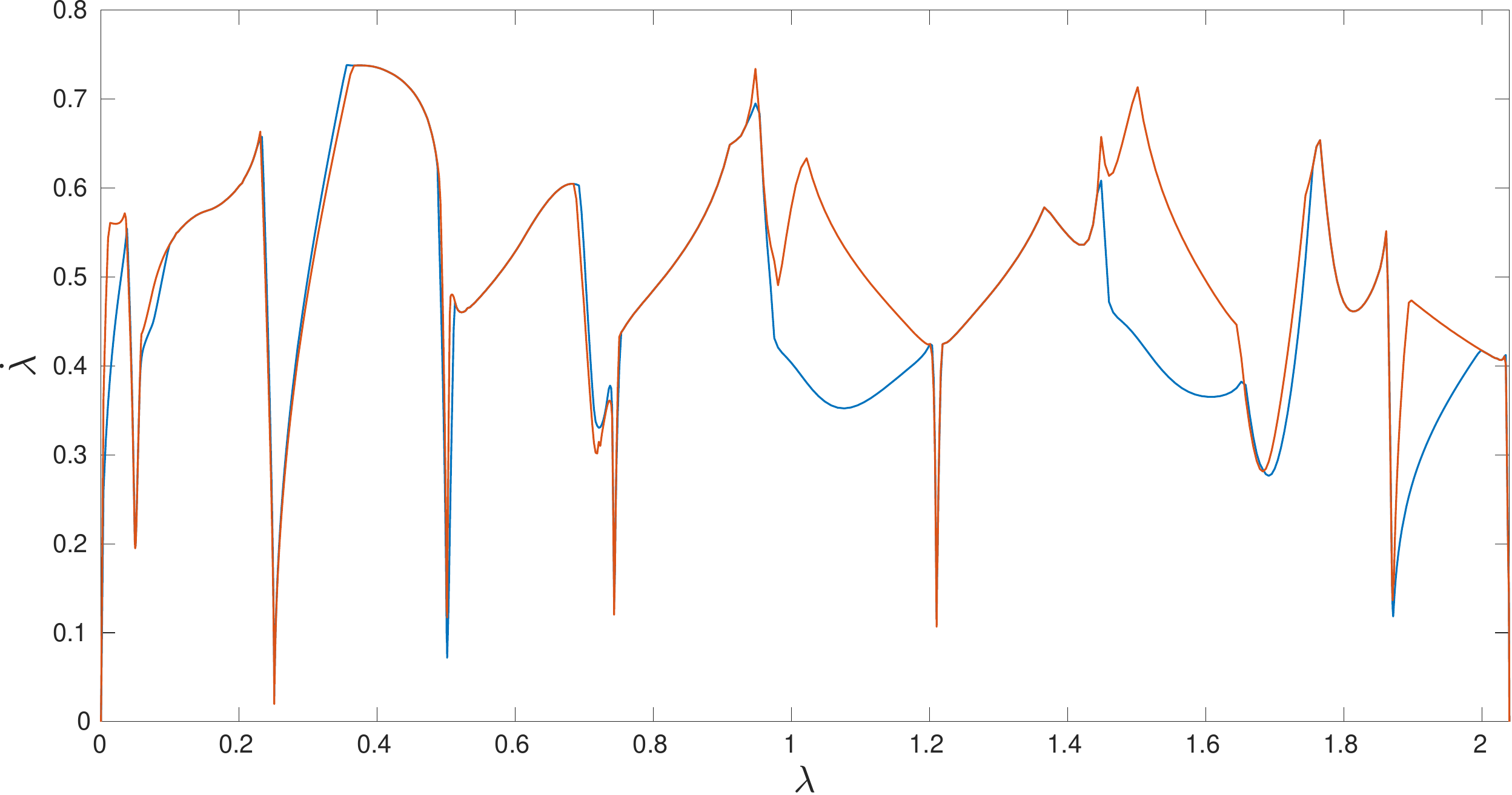}
\caption{Phase plane trajectory on the $(\lambda \mdash \dot \lambda)$ plane. In blue, the external wrenches are considered; in orange, they are neglected}
\label{fig:PPT}
\end{figure}

The path showed in \figref{fig:ACG} is time-parametrized using TOTP with DP. The grid dimensions are set to $N_\lambda=500$ and $N_{\dot \lambda}=5000$, with $\dot \lambda \in [0; 1.0]$. The Phase plane trajectory (PPT), as resulted from planning, is plotted (in blue) in \figref{fig:PPT}, and compared with the PPT planned excluding interaction wrenches in the dynamic model (in orange). In TOTP, the trajectory tracking time is inversely proportional to the area beneath the PPT \cite{shin_minimum-time_1985}. It is immediate to notice that the action of exerting wrenches on the blackboard has a global effect of lowering the PPT, resulting in a slower motion.

In terms of the solution to the optimization problem, the minimum of the cost function (\ie, the time, according to \eqref{eq:t_opt}) found by the algorithm described in Section \ref{sec:time_optimal_trajectory_planning_with_interaction_with_the_enviornment} is 4.51 seconds, while adopting the state-of-the-art DP algorithm gives an optimal time of 4.07 seconds, thus yielding a faster (and, if interaction does happen during the execution, practically unfeasible) trajectory. As the optimization problem remains the same, employing other compatible state-of-the-art optimization techniques should produce a similar time.

Despite the global effect, it is interesting to notice that, locally, for some $\lambda$, the PPT with interaction wrenches is higher than the contact-free one. In those regions, the manipulator is gaining dynamic capabilities in exerting wrenches on the environment. This fact is more easily analyzed in the torque domain.

\begin{figure}[t] 
    \centering
  \subfloat[base joint\label{fig:planned_torques_1_base}]{%
       \includegraphics[width=\columnwidth]{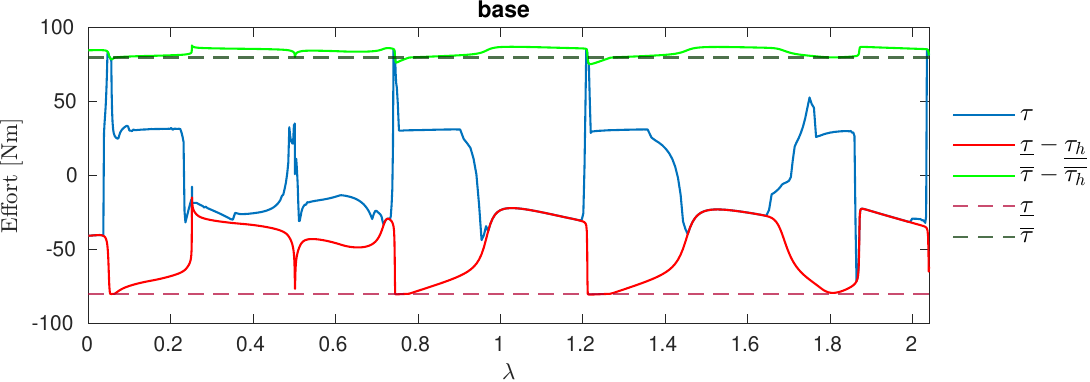}}
    \\
  \subfloat[shoulder joint\label{fig:planned_torques_2_shoulder}]{%
        \includegraphics[width=\columnwidth]{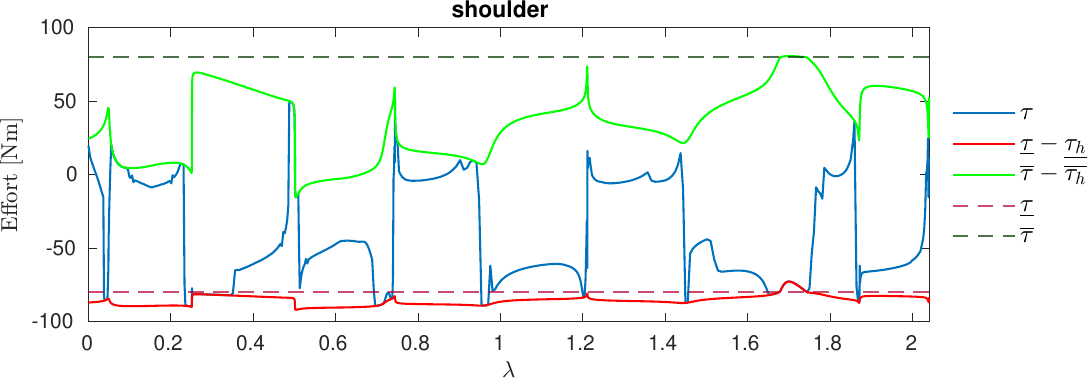}}
  \caption{Planned torques accounting for exerted wrenches and (modified) limits for first (a) and second (b) joint}
  \label{fig:planned_torques} 
\end{figure}

The effect of \eqref{eq:l_j} and \eqref{eq:u_j} is to impose constraints on $\ddot \lambda$, in fact changing the torque limits $\underline{\bm \tau}$ and $\overline{\bm \tau}$ by path-dependent offsets $\underline{\bm \tau_h}(\lambda)$ and $\overline{\bm \tau_h}(\lambda)$. The plots in \figref{fig:planned_torques} prove that the planned torques do not break the modified limits and saturate over them when the joint velocities do not saturate. Yet, it is possible that the available torques are beyond $\overline{\bm \tau}$ (or $\underline{\bm \tau}$), with $\overline{\bm \tau} - \overline{\bm \tau_h}(\lambda)$ (or $\underline{\bm \tau} - \underline{\bm \tau_h}(\lambda)$) being less strict than the technological torque limits. For example, this is the case of hard acceleration/deceleration phases where the torque needed at joint $j$ to exert the tangential force compensates the term $a_j(\lambda) \ddot{\lambda}$, indeed increasing the residual torque, to be employed for time minimization. Conversely, the most common effect is that the tangential force produces, at joint $j$, a torque that opposes to the accelerating/decelerating one, in fact limiting the dynamic capabilities of the manipulator. In \figref{fig:planned_torques}, only base and shoulder joints are shown as they are the limiting joints for the ``ACG'' path. The other joints adapt to allow for trajectory tracking.

\begin{figure}[t] 
    \centering
  \subfloat[base joint\label{fig:planned_torques_1_base_no_forces}]{%
       \includegraphics[width=\columnwidth]{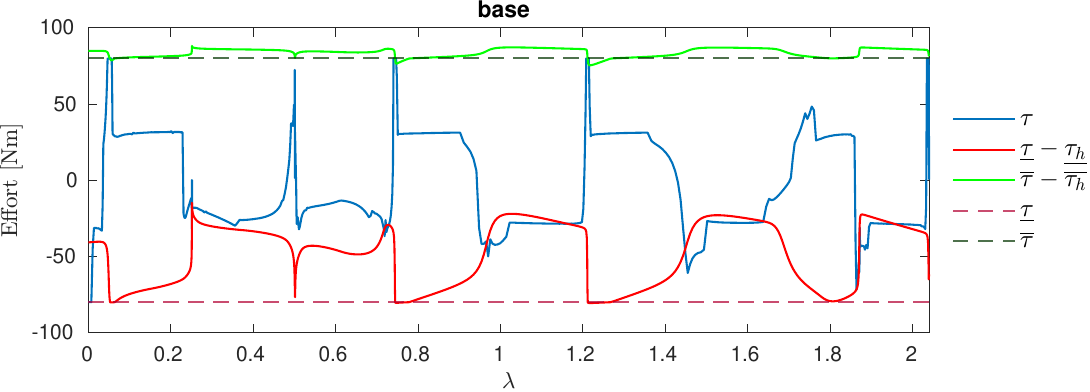}}
    \\
  \subfloat[shoulder joint\label{fig:planned_torques_2_shoulder_no_forces}]{%
        \includegraphics[width=\columnwidth]{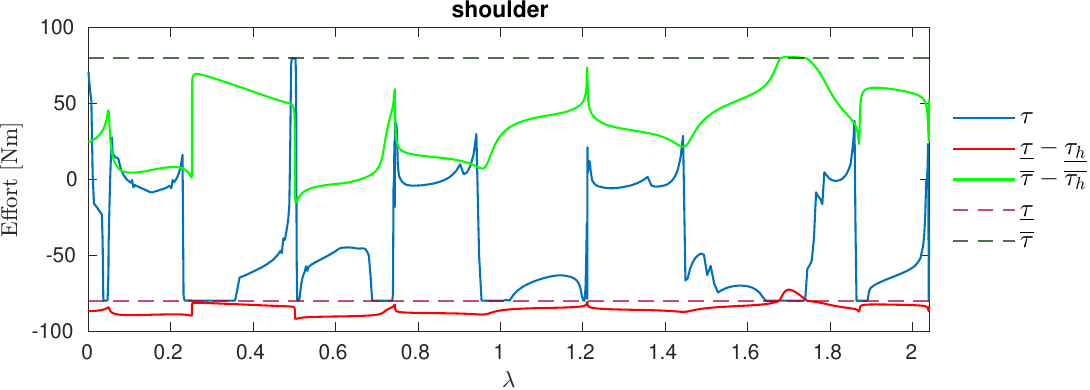}}
  \caption{Planned torques ignoring exerted wrenches and (modified) limits for first (a) and second (b) joint}
  \label{fig:planned_torques_no_forces} 
\end{figure}

In contrast, Figure \ref{fig:planned_torques_no_forces} shows the torques planned without including external wrenches in the algorithm, thus proving the effectiveness of adopting the proposed framework. It is immediate to notice that, for some joint $j$, the planned torque can break the (modified) limit in certain portions of the path. For example, in Figure \ref{fig:planned_torques_1_base_no_forces}, $\tau$ overcomes the lower bound $\underline \tau - \underline{\tau_h}$ for $\lambda \in [1; 1.2]$. This phenomenon has a critical practical outcome, causing a possible failure at execution. In fact, if the robot does exert wrenches, then the torques actually available for joint actuation would not be sufficient for performing the motion as it was planned, resulting in a path deviation.

\begin{figure}[!b]
\centering
\includegraphics[width=\columnwidth]{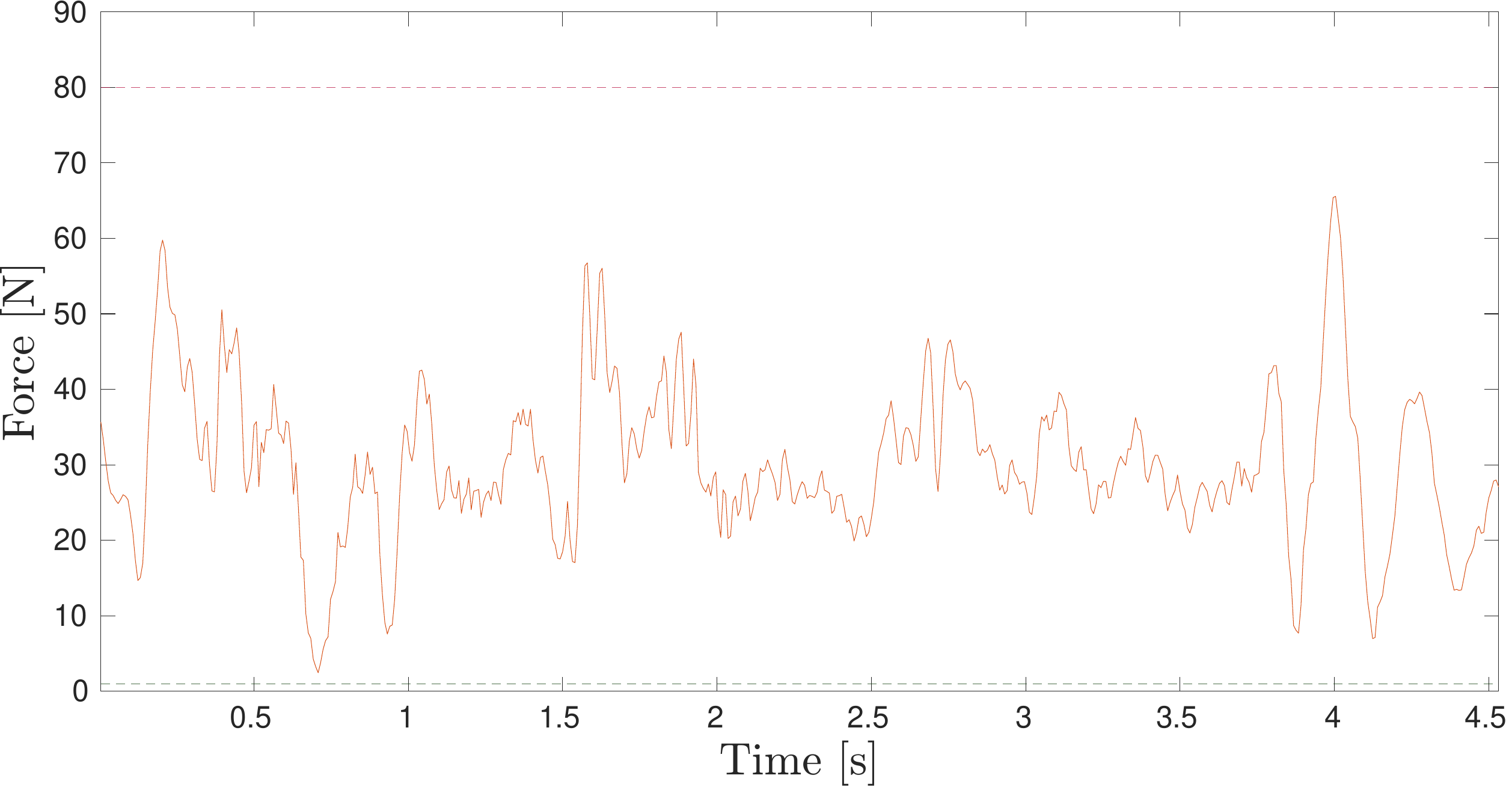}
\caption{Contact force along the normal direction (solid line) during the execution of the time-optimal planned trajectory and contact force limits (dashed lines)}
\label{fig:forces}
\end{figure}

The admittance controller, with the selected tuning, guarantees that the measured force is within bounds, as evident from \figref{fig:forces}, confirming that the execution respects the task specifications considered at planning level. As a consequence, the chalk does not get damaged, but a clear stroke can be delivered.

In order to stress the necessity of explicitly adding the wrench bounds in the planning process, it is worth recalling that only with this inclusion the robot's joint torques and velocities are guaranteed to respect the manipulator's physical capabilities, accounting for both motion and interaction efforts. On the contrary, if interaction wrenches are disregarded in planning, then the only solution for a safe task execution is to down-scale the PPT, therefore manually lowering the trajectory's speed. Clearly, this approach would yield the following drawbacks:
\begin{enumerate}
\item the planning process will no longer be \textit{optimal} with respect to the actual manipulator’s capabilities;
\item the time parametrization will no longer benefit from the less strict joint torque limits resulting from the interaction, as previously discussed in this section.
\end{enumerate}

\begin{figure}[!t]
\centering
\includegraphics[width=\columnwidth]{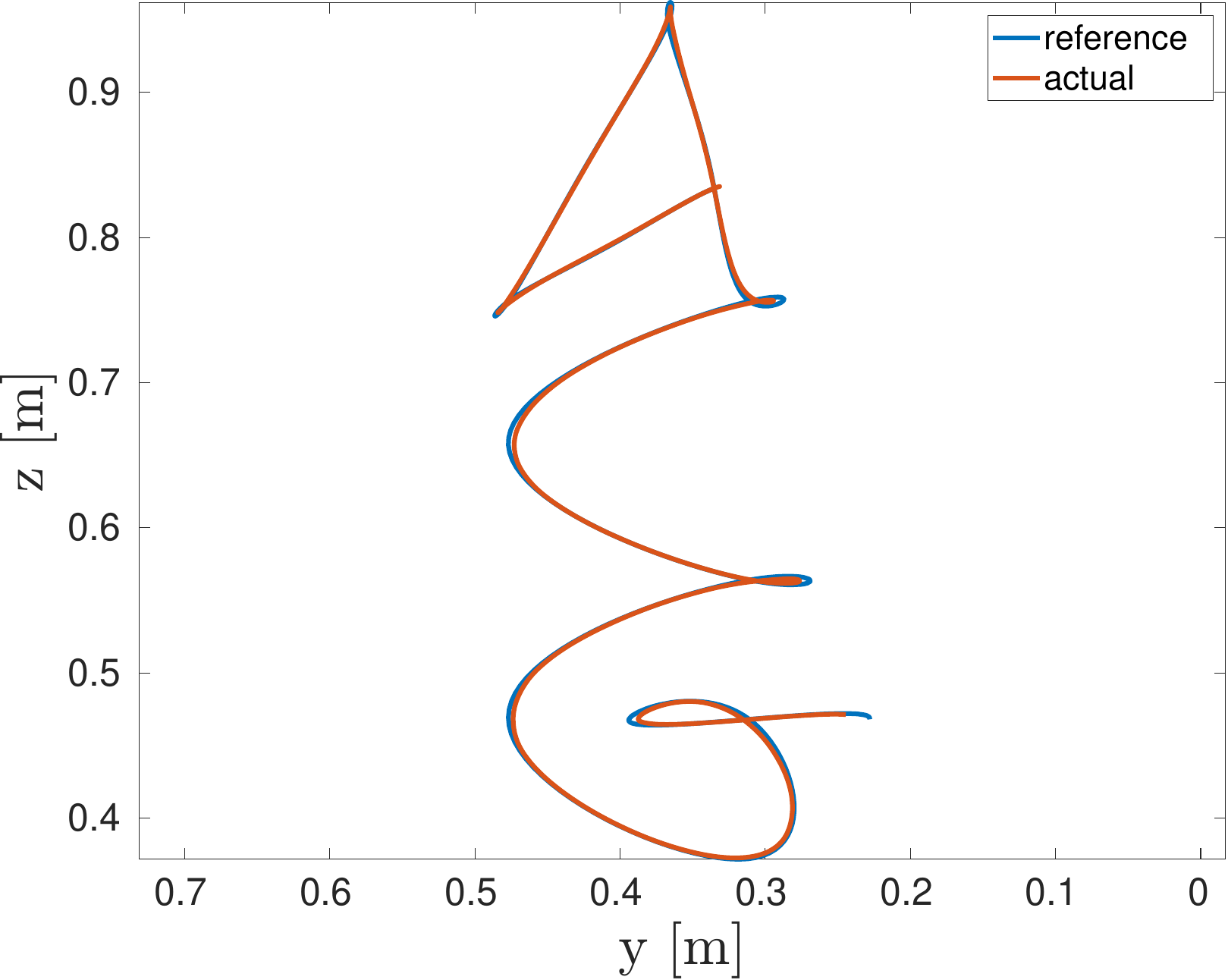}
\caption{Comparison between reference and tracked paths after the trajectory execution}
\label{fig:tracking}
\end{figure}

\figref{fig:tracking} provides a qualitative illustration of the trajectory tracking performance. The path is accurately tracked, considering the type of task (writing is effective as long as letters can be distinguished from one another), with the higher  deviations arising at the segments of highest curvature. This is also where the robot hardly accelerates and decelerates, while coping with external forces. It is anyhow worth remarking that the actual trajectory and force tracking performances depend on the controller adopted in the execution phase, as also noted in similar works \cite{kaserer_time_2020}. For example, trajectory tracking performances can be improved by making the robot stiffer along the motion/writing plane, but improving the controller performance goes beyond the scope of this communication.

\section{Conclusion} \label{sec:conclusions}

This paper deals with optimal trajectory planning for robotic manipulators with interaction with an unmodeled environment.

The proposed strategy to consider interaction is imposing limits on the wrenches to exert on the environment, also accounting for uncertainties on their dynamic characterization. In this communication, for simplicity, limits are imposed on linear forces only, but the same approach can be easily extended to momenta.

This methodology is compatible with every implementation of the planning algorithm and additional joint space and task space constraints can be included without any modification, regardless of the performance index to optimize.

The validation experiment consists in a time-optimal planned trajectory executed on a real UR10 robot, employing an admittance controller to deal with the interaction at control level. Indeed, the performances are only limited by the capabilities of the controller to track forces and positions accurately, while moving at very high speeds. The inclusion of wrenches at planning level affects the trajectory tracking time: while it is intuitive to expect that interaction slows motion down, in some cases, interaction can be beneficial to motion, indeed improving the dynamic capabilities of the robot.

Restricting the wrench limits requires the interaction controller to accurately track the wrench reference. It would be interesting, in future works, to assess the capabilities of a different control strategy, e.g., direct force control, to accurately track the force reference while coping with the high-speed dynamics of time-optimal motion. Also, controllers introduce their own dynamics which could be considered at planning level.

\addtolength{\textheight}{-12cm}   % This command serves to balance the column lengths

%%%%%%%%%%%%%%%%%%%%%%%%%%%%%%%%%%%%%%%%%%%%%%%%%%%%%%%%%%%%%%%%%%%%%%%%%%%%%%%%

%%%%%%%%%%%%%%%%%%%%%%%%%%%%%%%%%%%%%%%%%%%%%%%%%%%%%%%%%%%%%%%%%%%%%%%%%%%%%%%%

%%%%%%%%%%%%%%%%%%%%%%%%%%%%%%%%%%%%%%%%%%%%%%%%%%%%%%%%%%%%%%%%%%%%%%%%%%%%%%%%

%%%%%%%%%%%%%%%%%%%%%%%%%%%%%%%%%%%%%%%%%%%%%%%%%%%%%%%%%%%%%%%%%%%%%%%%%%%%%%%%

\bibliographystyle{IEEEtran}
\bibliography{iros-2022-totp-interaction}

\end{document}